\documentclass[conference]{IEEEtran}
\usepackage{cite}
\usepackage{amsmath,amssymb,amsfonts}
\usepackage{algorithmic}
\usepackage{graphicx}
\usepackage{textcomp}
\usepackage{xcolor}
\usepackage{hyperref}
\usepackage{booktabs}

\begin{document}

\title{Breaking the Geometric Bottleneck: Contrastive Expansion in Asymmetric Cross-Modal Distillation}

\author{\IEEEauthorblockN{Kabir Thayani}
\IEEEauthorblockA{\textit{Independent Researcher} \\
India \\
thayanikabir.official@gmail.com}
}

\maketitle

\begin{abstract}
Knowledge distillation between asymmetric architectures often induces severe geometric constraints on the learned representation space. In this work, we investigate the dimensional collapse phenomenon when distilling global Vision Transformers into strictly capacity-constrained, local-receptive-field CNNs (0.5M to 8.0M parameters). Using strictly centered Singular Value Decomposition (SVD) and Shannon Entropy Effective Rank, we confirm a capacity-agnostic collapse under standard cosine distillation: a CLIP ViT-B/32 Teacher exhibits an Effective Rank of 88.68 on CIFAR-10, while all cosine-distilled students collapse to $\sim$17, regardless of parameter count (0.5M--8.0M). An auxiliary InfoNCE objective expands this to $\sim$41 dimensions.

Critically, we go beyond spectral geometry to ask whether this expansion is functionally useful. Using multi-seed linear-probe evaluation, we find that InfoNCE expansion \textit{degrades} downstream classification accuracy by 15--18 percentage points relative to the collapsed baseline, despite more than doubling Effective Rank. A class-structure decomposition traces this to signal dilution: InfoNCE's class-blind uniformity pressure weakens class-discriminative structure in the original dimensions while adding only weakly class-relevant structure in the new ones. We then test a label-aware alternative, Supervised Contrastive distillation. On CIFAR-100, where we swept student capacity directly, its Effective Rank is \textit{invariant to model capacity} (constant at $\sim$22 dimensions across a 16$\times$ parameter range); on CIFAR-10, at a single tested width, it settles to a lower rank ($\sim$8) while matching baseline accuracy. Where we swept temperature instead of capacity (CIFAR-100), rank and downstream accuracy increased together monotonically across $\tau \in [0.05, 1.0]$. These results indicate that Effective Rank alone is not a reliable proxy for representation quality: whether expansion helps or harms downstream performance depends on whether the driving objective is label-aware, not on the magnitude of the expansion itself. We report all results with mean $\pm$ standard deviation over three seeds where computationally feasible, and explicitly note where evaluation depth differs between our CIFAR-10/CLIP and CIFAR-100/DINOv2 experiments.
\end{abstract}


\section{Introduction}
The deployment of state-of-the-art foundational vision models is heavily bottlenecked by their massive parameter counts. Knowledge Distillation \cite{hinton2015distilling} is the standard paradigm for compressing these models for edge deployment. However, transferring knowledge from a global-receptive-field Vision Transformer (ViT) into a strictly local-receptive-field Convolutional Neural Network (CNN) \cite{he2016deep} creates a severe asymmetric bottleneck.

Previous studies have shown that embedding spaces in deep neural networks often suffer from intrinsic anisotropy and dimensional collapse \cite{jing2022understanding}. A natural response, adopted in an earlier version of this work, is to treat collapse as unambiguously harmful and Effective Rank expansion as unambiguously beneficial. In this revision, we test that assumption directly against downstream task performance, and find it does not hold in general.

Our contributions are six-fold:
\begin{enumerate}
    \item We confirm, with multi-seed evidence, capacity-agnostic dimensional collapse under standard cosine distillation (CLIP $\to$ CNN, CIFAR-10): students collapse to Effective Rank $\sim$17 regardless of scale (0.5M--8.0M parameters).
    \item We isolate the cause via architectural ablation, confirming collapse is objective-induced rather than architecture-induced.
    \item We show that an auxiliary InfoNCE objective expands Effective Rank ($\sim$17 $\to$ $\sim$41 on CIFAR-10; $\sim$54 $\to$ $\sim$88 on CIFAR-100), \textbf{but} multi-seed linear-probe evaluation reveals this expansion \textit{reduces} downstream accuracy by 15--18 points on CIFAR-10 and $\sim$3.6 points on CIFAR-100.
    \item Using a class-structure decomposition (between-class variance per spectral dimension), we show this drop is explained by signal dilution: InfoNCE weakens class-discriminative structure in the originally-collapsed dimensions while adding only weakly-informative structure elsewhere.
    \item We test a label-aware alternative, Supervised Contrastive (SupCon) distillation. On CIFAR-100, where we directly swept student capacity, its Effective Rank is invariant to width (flat across a 16$\times$ parameter sweep); where we instead swept the contrastive temperature $\tau$ at fixed width, rank and downstream accuracy increased together monotonically across $\tau \in [0.05, 1.0]$.
    \item We synthesize these results into a revised claim: Effective Rank is not intrinsically a proxy for representation quality. Its relationship to downstream performance is objective-dependent -- expansion driven by a class-blind objective harms accuracy, while expansion driven by a class-aware objective's own operating point can track accuracy positively.
\end{enumerate}

\section{Methodology}

\subsection{Architectures and Distillation}
We designed a custom Scalable CNN Student architecture with a width expansion factor, yielding three variants: Student-S (0.5M), Student-M (2.0M), and Student-L (8.0M). We used two Teacher networks: CLIP ViT-B/32 (multi-modal contrastive pretraining, evaluated on CIFAR-10) and DINOv2 ViT-Base (pure vision self-supervision, evaluated on CIFAR-100).

The baseline objective is cosine distance distillation:
\begin{equation}
    \mathcal{L}_{cos} = \frac{1}{N} \sum_{i=1}^{N} \left( 1 - \frac{z_s^{(i)} \cdot z_t^{(i)}}{\|z_s^{(i)}\| \|z_t^{(i)}\|} \right)
\end{equation}

To counter dimensional collapse, we add an auxiliary InfoNCE loss across augmented student views:
\begin{equation}
    \mathcal{L}_{NCE} = -\log \frac{\exp(\text{sim}(z_{s1}, z_{s2})/\tau)}{\sum_{k} \exp(\text{sim}(z_{s1}, z_{sk})/\tau)}
\end{equation}

As a label-aware alternative, we introduce a Supervised Contrastive loss \cite{khosla2020supervised}, where positive pairs are same-class samples within the batch rather than augmented views of a single sample:
\begin{equation}
    \mathcal{L}_{SupCon} = -\frac{1}{|P(i)|}\sum_{p \in P(i)} \log \frac{\exp(z_i \cdot z_p / \tau)}{\sum_{a \neq i} \exp(z_i \cdot z_a / \tau)}
\end{equation}
where $P(i)$ is the set of same-class samples in the batch excluding $i$ itself. Final training losses are $\mathcal{L}_{cos} + 0.5\,\mathcal{L}_{NCE}$ or $\mathcal{L}_{cos} + 0.5\,\mathcal{L}_{SupCon}$.

\subsection{Rigorous Spectral Evaluation}
All embedding matrices were strictly centered prior to SVD: $Z_c = Z - \mu_Z$. Shannon Entropy Effective Rank uses the normalized squared singular values:
\begin{equation}
    p_i = \frac{\sigma_i^2}{\sum \sigma_j^2}, \quad ER = \exp\left(-\sum p_i \ln p_i\right)
\end{equation}

\subsection{Downstream Evaluation (New in this Revision)}
For every trained student, we fit a linear probe (logistic regression) on frozen train-split embeddings and report clean test-set accuracy. Where computationally feasible, we repeat training across 3 seeds (42, 123, 456) and report mean $\pm$ standard deviation. Noise-robustness curves use the same linear probe evaluated on test images perturbed with additive Gaussian noise at severities $\sigma \in \{0, 0.05, 0.1, 0.2\}$.

\subsection{Class-Structure Decomposition (New in this Revision)}
To determine whether spectral dimensions carry class-relevant information, we project centered embeddings onto their own principal components ($Z_{proj} = Z_c V^\top$, from the SVD $Z_c = U\Sigma V^\top$) and compute, per dimension $j$, the ratio of between-class variance to total variance:
\begin{equation}
    \text{Sep}_j = \frac{\frac{1}{N}\sum_{c} n_c (\bar{z}_{c,j} - \bar{z}_j)^2}{\text{Var}(z_{\cdot,j})}
\end{equation}
where $c$ indexes classes, $n_c$ is the class count, and $\bar{z}_{c,j}$ is the class-$c$ mean along dimension $j$. Higher values indicate the dimension separates classes well; values near zero indicate the dimension is dominated by within-class (instance-level) variation. When comparing dimension bands across models with different Effective Ranks, we report separability over each model's \textit{own} top-$k$ dimensions (with $k$ set to that model's measured Effective Rank) to avoid diluting the score with near-zero-variance tail dimensions -- an artifact we identified and correct for explicitly in Section~V.

\section{Results and Analysis (Phase 1: CLIP / CIFAR-10)}

\subsection{Capacity-Agnostic Dimensional Collapse}
The Teacher model exhibited an Effective Rank of 88.68; all cosine-distilled students collapsed to $\sim$17 regardless of scale (Table \ref{tab:metrics}), consistent with the original submission.

\begin{table}[h]
\centering
\caption{Baseline Scaling Metrics (CLIP / CIFAR-10)}
\label{tab:metrics}
\begin{tabular}{@{}lccc@{}}
\toprule
\textbf{Model} & \textbf{Params} & \textbf{Effective Rank} \\ \midrule
Teacher-CLIP   & 500M            & 88.68                   \\
Student-S      & $\sim$0.5M      & 15.91                   \\
Student-M      & $\sim$2.0M      & 16.62                   \\
Student-L      & $\sim$8.0M      & 16.66                   \\ \bottomrule
\end{tabular}
\end{table}

\subsection{Objective vs. Architecture Ablation}
An architectural ablation (strictly local CNN, semi-global Hybrid, global Tiny ViT) confirmed collapse is objective-induced: CNN 15.90, Hybrid 14.54, Tiny ViT 8.70. None escaped the ceiling, confirming the cosine objective -- not receptive field -- is the binding constraint.

\subsection{Contrastive Expansion Revisited: Rank vs. Downstream Accuracy}
\label{sec:phase1-revisited}
We repeated the InfoNCE expansion experiment (Student-M, width factor 2) across 3 seeds and, critically, added linear-probe evaluation alongside the spectral measurement. Table \ref{tab:phase1_multiseed} reports the result.

\begin{table}[h]
\centering
\caption{Multi-Seed Rank and Downstream Accuracy (CLIP / CIFAR-10, Student-M)}
\label{tab:phase1_multiseed}
\begin{tabular}{@{}lcc@{}}
\toprule
\textbf{Objective} & \textbf{Effective Rank} & \textbf{Probe Accuracy} \\ \midrule
Baseline (cosine only)      & $16.75 \pm 0.17$ & $72.24\% \pm 0.22\%$ \\
InfoNCE (cosine + NCE)      & $40.75 \pm 1.81$ & $54.15\% \pm 1.34\%$ \\
Supervised Contrastive      & $8.01 \pm 0.06$  & $71.66\% \pm 0.73\%$ \\ \bottomrule
\end{tabular}
\end{table}

This result revises our central claim from the original submission. InfoNCE more than doubles Effective Rank (16.75 $\to$ 40.75) but this comes with an \textbf{18.1 percentage point drop} in downstream accuracy, with non-overlapping confidence intervals across seeds -- a large, consistent, statistically robust effect, not noise. Meanwhile, Supervised Contrastive \textit{reduces} Effective Rank below baseline (to 8.01) while \textit{matching} baseline accuracy (71.66\% vs.\ 72.24\%). Across these three objectives, Effective Rank and downstream accuracy are not positively related -- if anything, the relationship is inverted.

\subsection{Class-Structure Decomposition (Phase 1)}
To explain the InfoNCE accuracy drop, we measured class-separability per spectral dimension (Table \ref{tab:sep_phase1}).

\begin{table}[h]
\centering
\caption{Class-Separability by Dimension Band (CLIP / CIFAR-10)}
\label{tab:sep_phase1}
\begin{tabular}{@{}lcc@{}}
\toprule
\textbf{Objective} & \textbf{Dims 1--16} & \textbf{Dims 17--38} \\ \midrule
Baseline (cosine only) & 0.2697 & 0.0128 \\
InfoNCE (cosine + NCE) & 0.0903 & 0.0212 \\ \bottomrule
\end{tabular}
\end{table}

InfoNCE does not simply append noisy dimensions on top of an intact baseline representation: it degrades separability in the \textit{original} leading dimensions by roughly 3$\times$ (0.2697 $\to$ 0.0903) while adding only weak signal in the new dimensions (0.0212, still far below either model's leading-dimension score). This is consistent with InfoNCE's instance-discrimination objective actively repelling same-class samples that cosine distillation had correctly clustered, since InfoNCE has no access to label information. We did not separately measure this decomposition for Supervised Contrastive in Phase 1; see Limitations.

\section{Generalization and Scaling (Phase 2: DINOv2 / CIFAR-100)}

\subsection{Inherited Anisotropy and Multi-Seed Confirmation}
The pure cosine baseline on CIFAR-100 achieved Effective Rank 54.49 $\pm$ 2.84 (3 seeds), substantially higher than CIFAR-10's $\sim$17, confirming that DINOv2's intrinsically more uniform geometry is partially inherited by the student even under a class-blind cosine objective. InfoNCE further expanded this to 88.14 $\pm$ 1.95.

\begin{table}[h]
\centering
\caption{Multi-Seed Rank and Downstream Accuracy (DINOv2 / CIFAR-100, Student-M)}
\label{tab:phase2_multiseed}
\begin{tabular}{@{}lcc@{}}
\toprule
\textbf{Objective} & \textbf{Effective Rank} & \textbf{Probe Accuracy} \\ \midrule
Baseline (cosine only) & $54.49 \pm 2.84$ & $40.69\% \pm 0.61\%$ \\
InfoNCE (cosine + NCE) & $88.14 \pm 1.95$ & $37.04\% \pm 0.56\%$ \\ \bottomrule
\end{tabular}
\end{table}

The InfoNCE accuracy cost is smaller in absolute terms on CIFAR-100 ($-3.65$ points) than on CIFAR-10 ($-18.1$ points), but is directionally identical and statistically consistent (non-overlapping std ranges). This replicates the Phase 1 finding on an independent teacher and dataset.

\subsection{Class-Structure Decomposition (Phase 2)}
\begin{table}[h]
\centering
\caption{Class-Separability by Dimension Band (DINOv2 / CIFAR-100)}
\label{tab:sep_phase2}
\begin{tabular}{@{}lcc@{}}
\toprule
\textbf{Objective} & \textbf{Dims 1--54} & \textbf{Dims 54--88} \\ \midrule
Baseline (cosine only) & 0.2775 & --- \\
InfoNCE (cosine + NCE) & 0.1825 & 0.0681 \\ \bottomrule
\end{tabular}
\end{table}

The same dilution pattern replicates: InfoNCE's own leading dimensions carry weaker class signal than baseline's (0.1825 vs.\ 0.2775), and its additional dimensions (0.0681) carry less signal still (ratio 0.373 relative to its own leading band).

\subsection{Supervised Contrastive: Capacity Invariance}
We swept Supervised Contrastive across the full student width range (0.5M--8.0M parameters, 16$\times$ range) to test whether the accuracy gap relative to baseline (40.69\% vs.\ $\sim$28.6\%, see Table \ref{tab:supcon_width}) reflects a capacity limitation.

\begin{table}[h]
\centering
\caption{Supervised Contrastive: Capacity Sweep (DINOv2 / CIFAR-100, $\tau=0.1$)}
\label{tab:supcon_width}
\begin{tabular}{@{}lcc@{}}
\toprule
\textbf{Model} & \textbf{Effective Rank} & \textbf{Probe Accuracy} \\ \midrule
SupCon-S (0.5M) & $22.30 \pm 0.80$ & $28.71\% \pm 1.05\%$ \\
SupCon-M (2.0M) & $21.69 \pm 0.09$ & $28.40\% \pm 0.39\%$ \\
SupCon-L (8.0M) & $21.61 \pm 0.78$ & $28.67\% \pm 0.87\%$ \\ \bottomrule
\end{tabular}
\end{table}

Effective Rank and accuracy are both flat across a 16$\times$ parameter range, ruling out capacity as the limiting factor. A fair per-dimension separability check (each model's own top-$k$ dimensions, $k \approx 22$) gave 0.2722, 0.2631, and 0.2579 respectively for S/M/L -- comparable to baseline's 0.2775 over its own top 54 dimensions. This indicates Supervised Contrastive's $\sim$22-dimensional representation is of comparable per-dimension quality to baseline's 54-dimensional one; the accuracy gap is a capacity-\textit{of-representation} limit (too few usable dimensions for 100 classes), not a signal-quality limit.

\subsection{Supervised Contrastive: Temperature Control}
\begin{figure}[h]
    \centering
    \includegraphics[width=\linewidth]{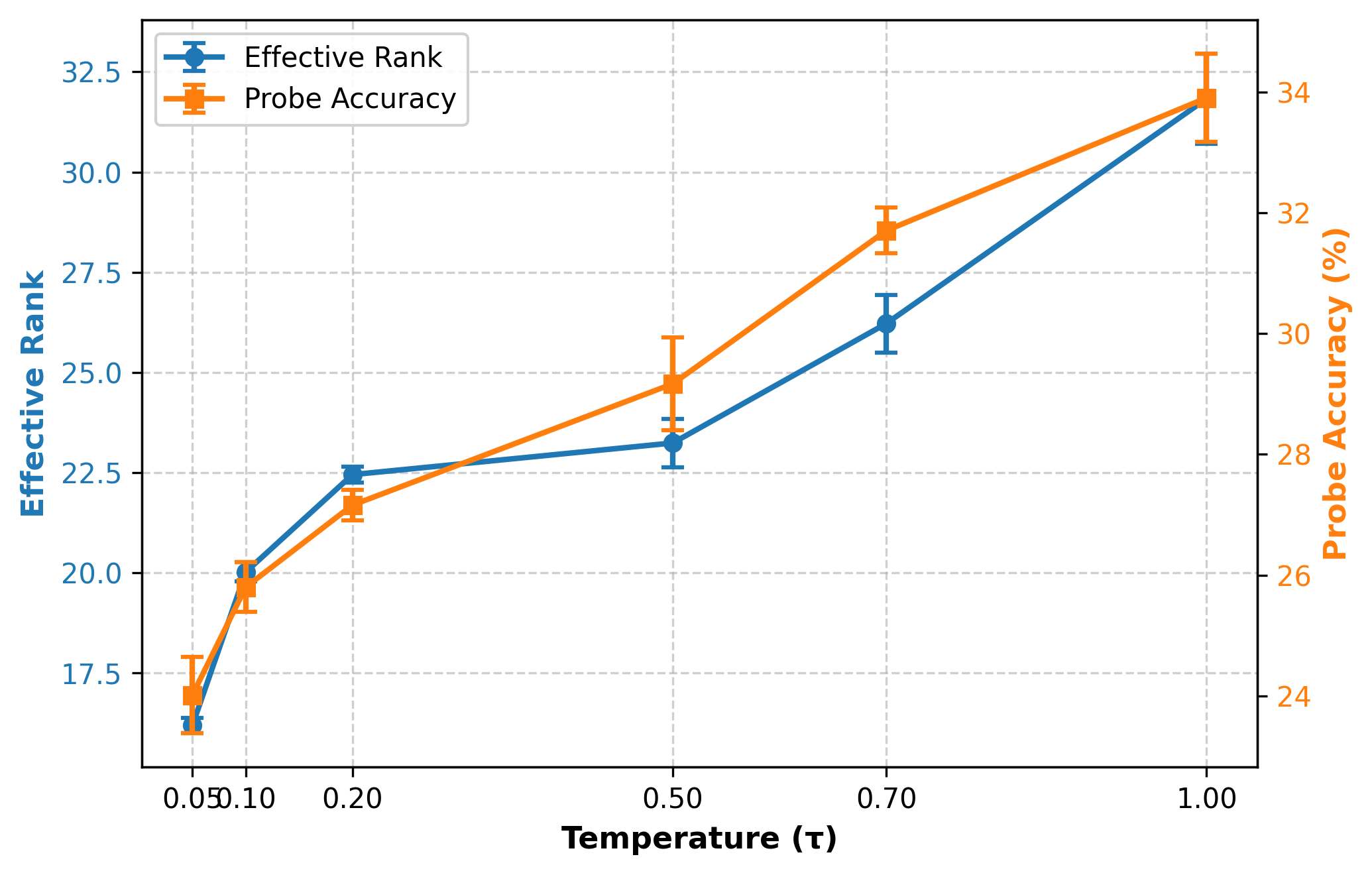}
    \caption{Supervised Contrastive: Effective Rank and downstream accuracy both increase monotonically with temperature $\tau$ (DINOv2 / CIFAR-100, Student-M, 3 seeds per point).}
    \label{fig:temp_sweep}
\end{figure}

Having ruled out model capacity, we swept the SupCon temperature $\tau \in \{0.05, 0.1, 0.2, 0.5, 0.7, 1.0\}$ at fixed width (Student-M), 3 seeds per setting (Table \ref{tab:supcon_temp}, Figure \ref{fig:temp_sweep}).

\begin{table}[h]
\centering
\caption{Supervised Contrastive: Temperature Sweep (DINOv2 / CIFAR-100, Student-M)}
\label{tab:supcon_temp}
\begin{tabular}{@{}lcc@{}}
\toprule
\textbf{$\tau$} & \textbf{Effective Rank} & \textbf{Probe Accuracy} \\ \midrule
0.05 & $16.19 \pm 0.19$ & $24.01\% \pm 0.63\%$ \\
0.10 & $20.02 \pm 0.24$ & $25.80\% \pm 0.41\%$ \\
0.20 & $22.45 \pm 0.20$ & $27.16\% \pm 0.25\%$ \\
0.50 & $23.24 \pm 0.60$ & $29.17\% \pm 0.77\%$ \\
0.70 & $26.22 \pm 0.72$ & $31.71\% \pm 0.38\%$ \\
1.00 & $31.83 \pm 1.12$ & $33.91\% \pm 0.73\%$ \\ \bottomrule
\end{tabular}
\end{table}

Unlike the capacity sweep, temperature produces a clean, monotonic increase in \textit{both} Effective Rank and downstream accuracy across the full range tested, with non-overlapping confidence intervals at every step. This is the one setting in this study where Effective Rank expansion and downstream accuracy improvement track together. We interpret this as follows: at low temperature, SupCon's clustering pressure is aggressive enough to over-compress the 100-class structure into too few usable dimensions; raising temperature relaxes this pressure, allowing the representation to recover class-relevant capacity that low-temperature training had discarded. We did not observe a plateau or reversal within the tested range ($\tau \le 1.0$); the curve's behavior beyond this point is untested (see Limitations).

\subsection{Noise Robustness}
As in the original submission, we evaluated linear-probe accuracy under additive Gaussian noise ($\sigma \in \{0, 0.05, 0.1, 0.2\}$). Consistent with the original finding, larger InfoNCE students showed reduced robustness at high noise severity relative to smaller ones (21.32\% vs.\ 17.03\% at $\sigma=0.2$, Student-S vs.\ Student-L). We note this robustness evaluation was conducted with a single seed per configuration in both the original submission and this revision, and should be treated with correspondingly lower confidence than the multi-seed rank/accuracy results reported above; we did not have the opportunity to extend it to 3 seeds in this revision cycle (see Limitations).

\section{Discussion: Effective Rank Is Not a Sufficient Proxy}
Synthesizing Sections III and IV, we revise our central claim. Effective Rank measures spectral spread, not task-relevant structure, and its relationship to downstream performance depends entirely on what is driving the expansion. This is consistent with the alignment-uniformity framework of \cite{wang2020understanding}: contrastive objectives trade off pulling positive pairs together (alignment) against spreading the representation uniformly over the embedding space (uniformity). InfoNCE's positives are augmented views of the same instance, so its uniformity pressure is applied without regard to class identity; Supervised Contrastive's positives are same-class samples, so its alignment term directly enforces class-relevant structure even as its uniformity term expands the space. Our temperature sweep (Section IV-D) can be read as moving along this alignment-uniformity trade-off within a single class-aware objective, which is why -- unlike InfoNCE's class-blind expansion -- it does not sacrifice downstream accuracy for rank:
\begin{itemize}
    \item \textbf{Class-blind expansion (InfoNCE)} consistently \textit{harms} downstream accuracy across two independent teacher/dataset pairs (CLIP/CIFAR-10, DINOv2/CIFAR-100), because its uniformity pressure does not distinguish same-class from different-class samples, diluting pre-existing class structure faster than it adds new structure.
    \item \textbf{Class-aware expansion (Supervised Contrastive)} shows no such penalty at its natural operating point (Phase 1: matches baseline accuracy at \textit{lower} rank than baseline) and, when its rank is deliberately varied via temperature rather than capacity, rank and accuracy move together positively (Phase 2).
\end{itemize}
The original submission's framing -- that dimensional collapse is a pathology to be reversed, and that Effective Rank expansion is inherently a sign of a healthier representation -- is not supported once downstream task performance is measured directly. A more accurate framing is that cosine distillation's collapse is, at least in part, a \textit{correct} compression onto class-relevant structure, and that objectives which expand the representation without reference to class identity can actively undo this benefit.

\section{Limitations}
This revision substantially extends the original submission's evaluation, but several gaps remain:
\begin{itemize}
    \item Noise-robustness results (Section IV-E) remain single-seed, unlike the rank/accuracy results reported elsewhere in this revision, and should be read with correspondingly lower confidence.
    \item The class-structure decomposition (Section III-D) was not performed for Supervised Contrastive in Phase 1; we infer high per-dimension signal quality from its accuracy parity with baseline, but this was not measured directly as it was in Phase 2.
    \item The temperature sweep (Section IV-D) shows a monotonic increase in both rank and accuracy up to $\tau=1.0$ with no observed plateau; whether this trend continues, plateaus, or reverses at higher temperatures is untested.
    \item All experiments use a single student architecture family (a width-scaled CNN) and two teacher/dataset pairs; generalization to other student architectures or additional datasets is untested.
    \item Linear-probe accuracy is used throughout as the downstream proxy; we do not evaluate fine-tuned transfer performance, which may behave differently.
\end{itemize}

\section{Conclusion}
We revisit the claim that dimensional collapse under cross-modal distillation is a pathology best reversed by contrastive expansion. Multi-seed downstream evaluation shows that InfoNCE-driven expansion, while doubling Effective Rank, substantially reduces linear-probe accuracy on two independent teacher/dataset pairs -- a result traced via class-structure decomposition to dilution of class-discriminative signal by a class-blind objective. A label-aware alternative, Supervised Contrastive distillation, avoids this penalty: its Effective Rank is invariant to model capacity but directly controlled by temperature, with rank and accuracy increasing together across the tested range. We conclude that Effective Rank, on its own, is not a reliable indicator of representation quality for downstream classification; whether an expansion is beneficial depends on whether the objective driving it has access to class information, not on the magnitude of the expansion itself.

\end{document}